\documentclass{article}
\usepackage{amssymb}
\usepackage{xcolor}
\definecolor{myorange}{RGB}{255,128,0}
\usepackage{algorithm}
\usepackage{algorithmicx}
\usepackage{algpseudocode}
\usepackage{microtype}
\usepackage{graphicx}
\expandafter\def\csname ver@subfig.sty\endcsname{}
\usepackage{booktabs} %
\usepackage{float}
\usepackage{bigstrut}
\usepackage{amsmath}
\usepackage{amssymb}
\usepackage{mathtools}
\usepackage{amsthm}
\usepackage{mathrsfs}
\usepackage{nicefrac}
\usepackage{dsfont}
\usepackage{enumitem}
\usepackage{subcaption}
\usepackage{graphicx,subfig}
\usepackage{float}
\usepackage[utf8]{inputenc} %
\usepackage[T1]{fontenc}    %
\usepackage{hyperref}       %
\usepackage{url}            %
\usepackage{booktabs}       %
\usepackage{amsfonts}       %
\usepackage{nicefrac}       %
\usepackage{microtype}      %
\usepackage{graphicx}
\usepackage{amssymb}
\usepackage{fdsymbol}
\usepackage{wrapfig}
\usepackage{lipsum}
\usepackage{enumitem}
\usepackage{stackengine}
\usepackage[font=small,labelfont=bf]{caption}
\usepackage{color}
\usepackage{adjustbox}
\usepackage{wrapfig}
\usepackage[normalem]{ulem}
\usepackage{xcolor}
\usepackage{enumitem}
\usepackage{booktabs}
\usepackage{amsmath}
\usepackage{xcolor}
\usepackage{multirow}
\setitemize{label=\textbullet, leftmargin=*, nolistsep}
\usepackage{listings}
\usepackage{rotating}
\usepackage{makecell}
\usepackage[normalem]{ulem}
\usepackage{amsmath}
\usepackage[all]{hypcap}
\usepackage{xspace}
\newcommand{\algoname}{IVE\xspace}
\usepackage[most]{tcolorbox}
\definecolor{exback}{RGB}{249,250,252}
\definecolor{exframe}{RGB}{200,210,225}
\newtcolorbox{examplebox}{
  colback=exback,
  colframe=exframe,
  boxrule=0.6pt,
  arc=2pt,
  left=5pt,right=5pt,top=4pt,bottom=4pt,
  enhanced,
  breakable,
}

\newcommand{\topic}[1]{\vspace{1mm}\textbf{#1}. }

\newcommand{\Tref}[1]{Table~\ref{#1}}

\usepackage[final]{corl_2025} 

\title{Imagine, Verify, Execute: Memory-Guided Agentic Exploration with Vision-Language Models}

\author{
\begin{tabular}{c}
\textbf{\quad \quad \quad Seungjae Lee$^{a*}$, Daniel Ekpo$^{a*}$, Haowen Liu$^{a}$} \\
\textbf{\quad \quad \quad Furong Huang$^{a,b\dagger}$, Abhinav Shrivastava$^{a\dagger}$, Jia-Bin Huang$^{a\dagger}$}
\end{tabular}\\
$^a$ University of Maryland, College Park, 
$^b$ Capital One \\
{\small\{sjaelee, daniekpo, hwl, furongh, abhinav, jbhuang\}@umd.edu}\\
{\small $^*$ Equal contribution \quad $^\dagger$ Equal advising}
\\
Project Page: \href{https://ive-robot.github.io/}{\textcolor{myorange}{https://ive-robot.github.io/}}
}

\begin{document}
\maketitle


\begin{abstract}
    Exploration is essential for general-purpose robotic learning, especially in open-ended environments where dense rewards, explicit goals, or task-specific supervision are scarce. 
    Vision-language models (VLMs), with their semantic reasoning over objects, spatial relations, and potential outcomes, present a compelling foundation for generating high-level exploratory behaviors.
    However, their outputs are often ungrounded, making it difficult to determine whether imagined transitions are physically feasible or informative. 
    To bridge the gap between imagination and execution, we present \algoname (\textbf{I}magine, \textbf{V}erify, \textbf{E}xecute), an agentic exploration framework inspired by human curiosity.  
    Human exploration is often driven by the desire to discover novel scene configurations and to deepen understanding of the environment.
    Similarly, \algoname leverages VLMs to abstract RGB-D observations into semantic scene graphs, imagine novel scenes, predict their physical plausibility, and generate executable skill sequences through action tools. 
    We evaluate \algoname in both simulated and real-world tabletop environments. 
    The results show that \algoname enables more diverse and meaningful exploration than RL baselines, as evidenced by a 4.1 to 7.8× increase in the entropy of visited states. 
    Moreover, the collected experience supports downstream learning, producing policies that closely match or exceed the performance of those trained on human-collected demonstrations.
\end{abstract}



\begin{figure}[htbp]
    \centering
    \includegraphics[width=0.9\linewidth]{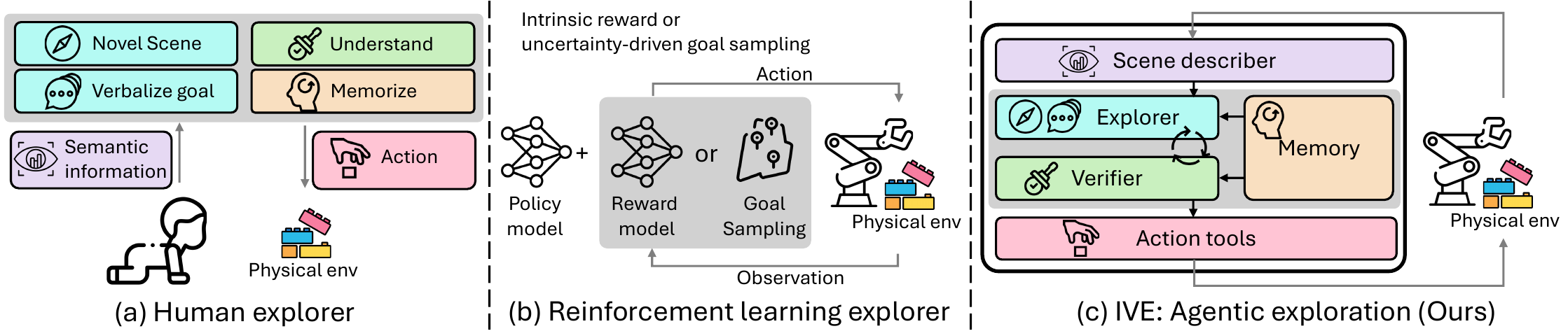}
    \caption{
    \textbf{Comparison of human, RL, and \algoname exploration strategies.} 
    \textbf{(a)} Humans explore by seeking novel scene configurations and understanding the environment \citep{ten2021humans, modirshanechi2023curiosity}, often enhanced by goal verbalization \citep{lidayan2025intrinsically}.
    \textbf{(b)} RL agents explore using a range of techniques, including intrinsic reward or goal sampling, to maximize the coverage of visited states.
    \textbf{(c)} \algoname (ours) leverages VLMs to structure exploration via scene description, exploration, verification, memory, and action tools, each aligned with key aspects of human exploration.
    }
    \vspace{-0.2in}
    \label{fig:main-diagram}
\end{figure}

\section{Introduction}

\begin{figure}[t!]
    \centering
    \includegraphics[width=0.9\linewidth]{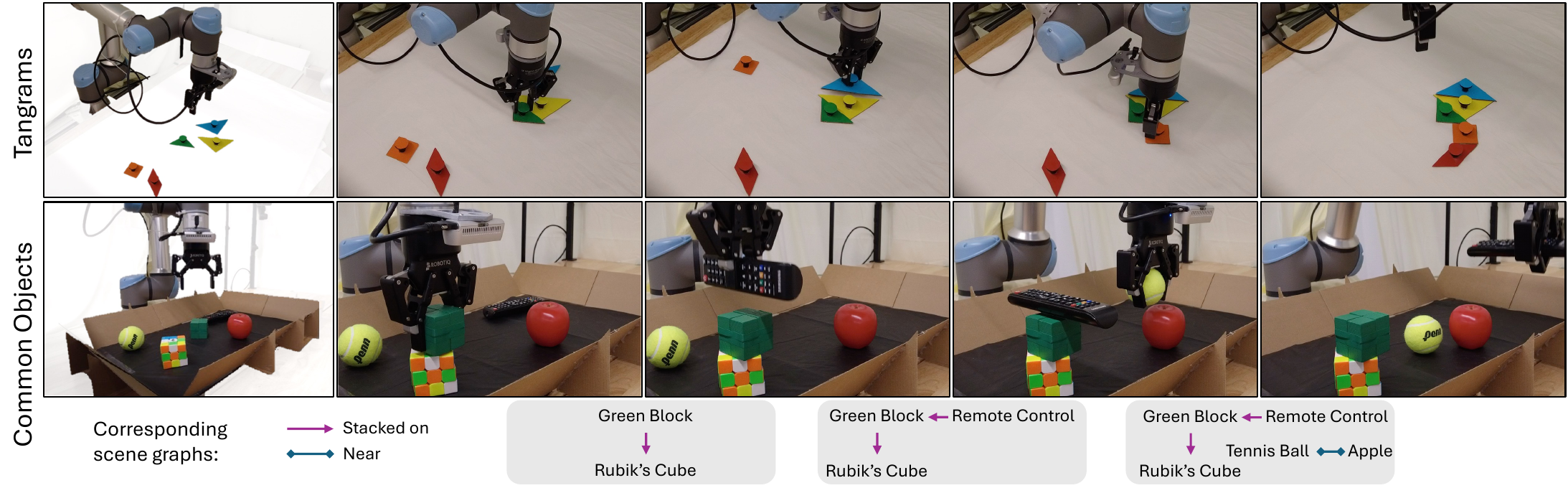}
    \caption{
    \textbf{Autonomous scene exploration with \algoname.}
    \algoname enables autonomous exploration of the scene with diverse objects, e.g., a tangram (\emph{top}) or common objects (\emph{bottom}).
    }
    \vspace{-5mm}
    \label{fig:autonomous_scene_explorer}
\end{figure}

Exploration is a fundamental capability for general-purpose robotic learning, particularly in open-ended environments where dense rewards, explicit goals, or demonstrations are scarce. 
In such settings, agents must autonomously discover diverse and meaningful interactions to support downstream learning and generalization~\citep{aubret2023information, lee2023cqm, sukhija2025maxinforl, gx2024efficient}. Reinforcement learning (RL) has been a dominant paradigm, often using intrinsic rewards to promote novelty~\citep{pathak2017curiosity, mendonca2021discovering,pmlr-v139-seo21a} or goal sampling to broaden state-space coverage~\citep{pong2019skew, hu2023planning}. While effective in simulation or low-dimensional tasks, RL methods often struggle in real-world robotic settings where environments are high-dimensional, semantically rich, and subject to physical constraints and safety risks. In these settings, the undirected or stochastic behaviors often encouraged by RL not only inefficient but also potentially hazardous.

Vision-language models (VLMs) offer a promising alternative. Their broad semantic knowledge and reasoning capabilities have enabled advances in robotic perception, reasoning, and high-level decision making~\citep{huang2023voxposer, shah2023lm, etukuru2024robot, tan2025mobile}. 
Building on these strengths, VLMs also offer a promising foundation for guiding exploration by generating hypothetical transitions, or quantifying novelty~\citep{kuang2024openfmnav, jiang2024multimodal, jiang2024roboexp, sancaktar2025sensei}.
However, their imagination often lacks grounding in physical dynamics: transitions may appear semantically plausible yet prove physically infeasible, redundant, or unsafe to execute~\citep{li2023can, hu2023look, elnoor2024robot}. 
Moreover, VLMs operate without structured memory of prior interactions, making it difficult to reason about which states have already been visited or which actions have been attempted.
This absence of memory and grounding often leads to redundant, implausible, or low-diversity generations that hinder effective exploration and downstream learning.

To address this challenge, we introduce \algoname (\textbf{I}magine, \textbf{V}erify, \textbf{E}xecute), a fully automated, VLM-guided system for agentic exploration, inspired by human exploration---by generating self-directed
goals, reacting to new information, and refining their understanding through experience \citep{ten2021humans, modirshanechi2023curiosity, lidayan2025intrinsically, gaven2025magellan}. 
\algoname enables agents to imagine novel future configurations, predict their feasibility based on recent interaction history, and execute selected behaviors via a library of skills. \algoname integrates imagination, verification, memory, and action execution 
into a closed loop, enabling exploration that is both semantically rich and physically grounded. The experience generated by \algoname is not only physically grounded and semantically rich, but also directly reusable for downstream policy learning and world model learning. In this work, we make the following key contributions:

\begin{itemize}
\item \textbf{Curiosity-Driven Exploration via Imagination and Verification.} We introduce \algoname, that combines memory-guided imagination with physical plausibility prediction to emulate human-like curiosity in embodied agents, achieving a 4.1 to 7.8× increase in state entropy over RL baselines.

\item \textbf{Reward-Free Data Collection with VLMs.} We develop a fully automated, vision-language model-guided agentic system for generating semantically meaningful interaction data—without requiring external rewards, demonstrations, or predefined goals, achieving 82\% to 122\% of the scene diversity exhibited by expert humans.

\item \textbf{Validation Across Downstream Tasks.} We provide extensive experiments in both simulated and real-world tabletop environments, demonstrating that \algoname improves exploration diversity and enables stronger policy learning and world model training compared to RL-based baselines.
\end{itemize}

\section{Related Work}

\topic{Exploration for Robotic Learning} 
Exploration is a fundamental challenge in enabling robots, and plays a critical role in building accurate models of environmental dynamics, discovering affordances, and identifying effective strategies for control. To tackle the challenge of exploration, many prior methods have often leveraged RL via intrinsic reward or goal sampling \citep{colas2022autotelic}. Intrinsic reward methods encourage exploration using prediction error \citep{pathak2017curiosity, burda2018exploration, mendonca2021discovering}, entropy \citep{pmlr-v139-seo21a, kim2023accelerating}, or state visitation \citep{martin2017count, shahidzadeh2023actexplore}, but often lack semantic understanding about the task, leading the agent to focus on perceptual novelty rather than task-relevant behaviors. In contrast, goal sampling methods guide exploration by sampling goals using temporal distance \citep{durugkar2021wasserstein, klissarov2023deep, bae2024tldr}, uncertainty \citep{cho2023outcome, hu2023planning}, or coverage \citep{pong2019skew, pitis2020maximum, yarats2021reinforcement, mahankali2024random}. While often more directed than intrinsic rewards, these methods still lack mechanisms to identify meaningful goals, and struggle in high-dimensional observation spaces, where learning a reliable latent representation or estimating uncertainty becomes challenging.

\topic{Vision-language models for Robotics}
VLMs have emerged as powerful tools for bridging visual perception and language-driven understanding. 
VLMs have demonstrated strong generalization across diverse tasks, enabling applications in robotic task generation~\citep{ahn2022can}, autonomous data collection~\citep{zhou2024autonomous}, evaluation~\citep{zhou2025autoeval}, and serving as high-level planners for low-level action tools~\citep{hu2023look, shah2024bumble}. 
While VLMs offer rich semantic understanding, leveraging them specifically for guiding exploration remains relatively underexplored. 
Recent efforts include using VLMs to improve goal-conditioned policies~\citep{zhou2024autonomous} or explore by ranking observations based on semantic interestingness~\citep{sancaktar2025sensei}. 
Similarly, we use a VLM in our work to guide exploration by prompting the VLM to imagine and propose physically plausible actions that will lead to novel scene configurations.

\topic{Scene graph for observation abstraction}
Scene graphs, which encode objects and their relationships as graph structures~\citep{johnson2015image}, emerged as a powerful tool for semantic scene understanding in computer vision~\citep{krishna2017visual, ji2020action, Johnson2018CVPRgeneration, Ashual2019SpecifyingOA, Dhamo2020SemanticIM, herzig2020learning}. 
Scene graphs have been integrated into robotic pipelines for grounding language instructions into executable actions \citep{rana2023sayplan, ni2024grid}, verifying plan feasibility~\citep{ekpo2024verigraph}, and supporting open-vocabulary understanding~\citep{gu2024conceptgraphs}.
Scene graph representation also enables hierarchical systems~\citep{ravichandran2022hierarchical, zhai2024sgbot, amiri2022reasoning}, highlighting their role in bridging high-level semantic reasoning with low-level physical execution.
Similar to prior methods~\citep{ekpo2024verigraph, rana2023sayplan}, we use scene graphs as an intermediate representation to the VLM. Beyond this role, we use scene graphs to measure the novelty of new scenes, encouraging the VLM to explore diverse scenes.

\section{Method}

\begin{figure}[t!]
    \centering
    \includegraphics[width=1\linewidth]{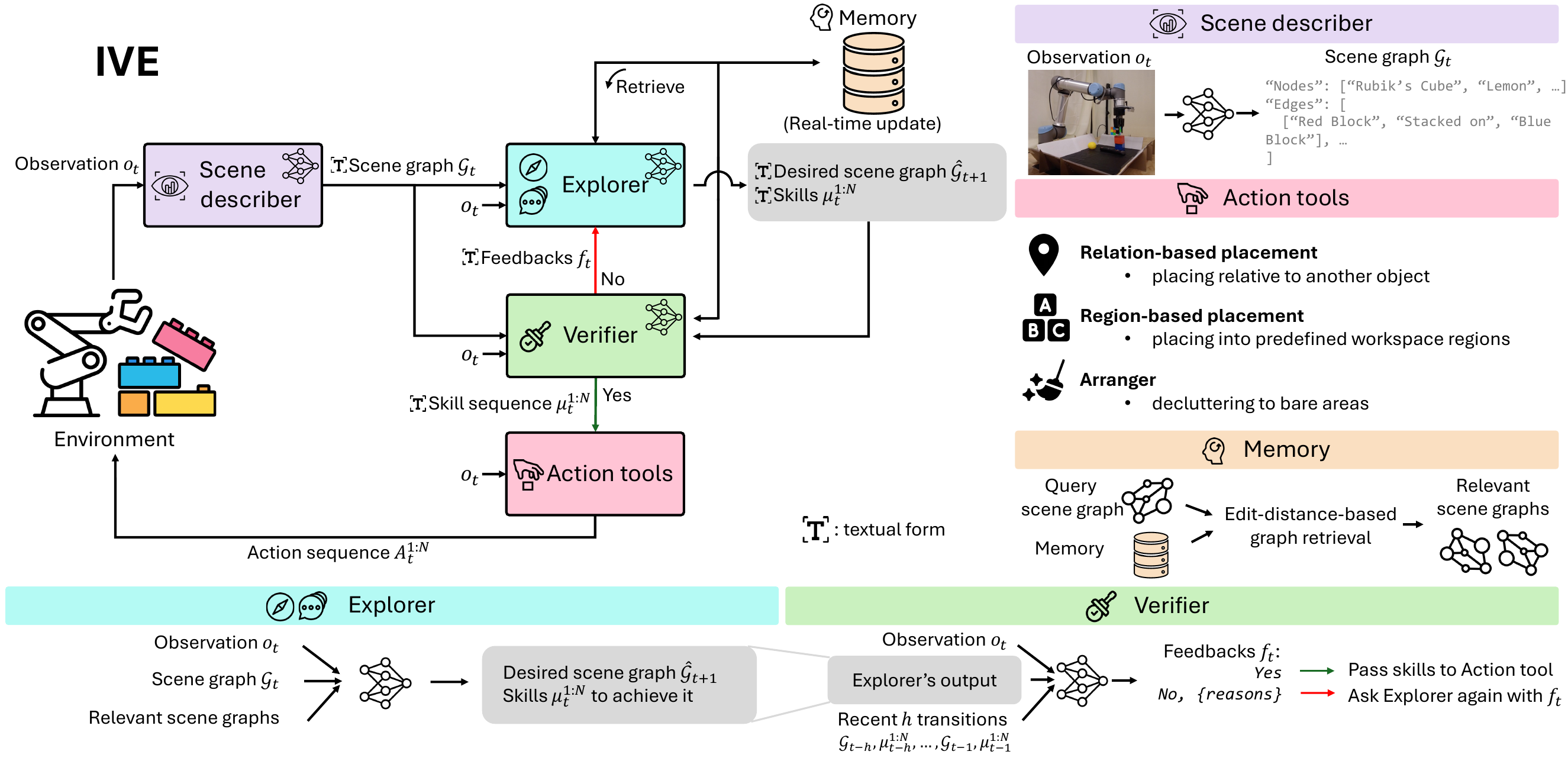}
        \caption{\textbf{Overview of \algoname.} 
        Given an observation $o_t$, the \textit{Scene Describer} constructs a semantic scene graph $\mathcal{G}_t$.
        The \textit{Explorer} leverages this representation, along with the current observation and retrieved past scene graphs, to generate (``imagine'') a candidate future scene graph $\hat{\mathcal{G}}_{t+1}$ and a sequence of skills $\mu_t^{1:N}$. 
        The \textit{Verifier} evaluates the feasibility of these imagined transitions using recent interaction history. 
        If verified, the skills are instantiated into low-level actions via the \textit{Action Tools} and executed by the robot. 
        Otherwise, the Explorer receives feedback and replans. The iterative process enables structured, curiosity-driven exploration grounded in semantic reasoning and informed by physical feasibility.
        }
        \vspace{-6mm}
    \label{fig:overview}
\end{figure}

\topic{Overview}
We propose a fully automated, VLM-driven
exploration system, \algoname, 
built on an agentic architecture composed of three \textbf{core} modules: 
the \textit{Scene Describer}, the \textit{Explorer}, and the \textit{Verifier}. 
(An overview of the \algoname system is shown in Figure~\ref{fig:overview}.)
\algoname begins by constructing an abstract, semantic representation of the current observation using the \textit{Scene Describer} (Section ~\ref{scene_describer}).
The \textit{Explorer} then proposes candidate future scenes along with skill sequences intended to achieve them (Section ~\ref{explorer}). 
These candidate plans are evaluated by the \textit{Verifier}, which predicts their physical plausibility and utility before execution (Section~\ref{verifier}). In addition to these \textbf{core} components, two \textbf{auxiliary} modules support the system: the \textit{Memory module} 
retrieves relevant past experiences to inform both the Explorer and Verifier (Section~\ref{memory}),
and the \textit{Action Tools module} translates the skill sequences into executable %
robot actions (Section~\ref{action_tool}).

\subsection{Scene Describer}\label{scene_describer} 

The Scene Describer module, powered by a VLM, constitutes the frontmost component of our system, abstracting raw observations into structured scene graphs that capture semantic object relationships. 
By converting high-dimensional visual inputs into compact, symbolic representations, this abstraction reduces the complexity of reasoning over raw sensory data, enabling more efficient exploration of novel scene configurations.

Given an observation $o_t$ (RGB image at timestep $t$),  the scene describer produces a scene graph $\mathcal{G}_t = (V, E)$, where $V$ denotes the set of objects (nodes) and $E$ encodes directed, typed semantic relations (edges) between pairs of objects.
Each object $v \in V$ refers to a unique object name, and edges $E$ include spatial and functional relations such as \texttt{Stacked on} and \texttt{Near}. The design of the scene graph can be adapted depending on the task, allowing flexibility in the level of abstraction and the types of relations captured.

\begin{examplebox}
\footnotesize
\textbf{Illustrative Example.}
Consider a tabletop scene containing a red cup, a blue block, and a tray. The Scene Describer may produce:
\[
\setlength{\abovedisplayskip}{4pt}
\setlength{\belowdisplayskip}{4pt}
\begin{aligned}
    V&=\{\text{Red cup}, \text{Blue block}, \text{Tray}, \ldots\}\\[-2pt]
    E&=\{(\text{Blue block},\text{Stacked on}, \text{Tray}), (\text{Red cup},\text{Near},\text{Tray})\}
\end{aligned}
\]
This abstract graph captures spatial relations without requiring dense 3D reconstruction. It allows the Explorer module to hypothesize meaningful future configurations (e.g., ``move the cup onto the tray'') while enabling the Verifier to assess feasibility (e.g., ``Is the tray already full?'') based on prior experiences.
\end{examplebox}

\subsection{Explorer} \label{explorer} 

The Explorer module is a VLM-based component that takes as input the current RGB observation $o_t$, the corresponding scene graph $\mathcal{G}_t$, and a set of relevant past experiences retrieved from memory (Figure \ref{fig:overview}). 
Using this contextual information, the Explorer imagines a future scene graph, $\hat{\mathcal{G}}_{t+1}$, that preserves the object set of $\mathcal{G}_t$ but alters the edge structure---representing new spatial or functional relationships among objects.
These imagined transitions enable the agent to reason about potential next states beyond those encountered previously.

To promote novelty and avoid redundancy, the Explorer compares $\hat{\mathcal{G}}_{t+1}$ with retrieved scene graphs from memory, encouraging transitions that are diverse and previously unseen. 
This memory-aware imagination supports a form of curiosity-driven exploration over structured symbolic representations.

In parallel with the imagined graph, the Explorer generates a sequence of $N$ high-level skills $\mu^{1:N}_t$ intended to transition the agent from $\mathcal{G}_t$ to $\hat{\mathcal{G}}_{t+1}$.
Each skill corresponds to a discrete, interpretable action primitive from a predefined skill library. These skill sequences are then passed to the Verifier for physical feasibility assessment \emph{prior} to execution.

\begin{examplebox}
\footnotesize
\textbf{Continuing Example.}
Returning to our earlier example, suppose the current scene graph encodes that the block is \texttt{Stacked on} the tray and the cup is \texttt{Near} the tray. The Explorer may propose a future graph where the cup is now \texttt{Stacked on} the tray and the block is \texttt{Near} the tray. It may then generate a skill sequence such as:
\[
\setlength{\abovedisplayskip}{4pt}
\setlength{\belowdisplayskip}{4pt}
\mu_t^{1:2} =\{\text{move}(\text{Red cup},\text{Stacked on}, \text{Tray}), \text{move}(\text{Blue block}, \text{To the left of}, \text{Tray})\}
\]
\end{examplebox}

\subsection{Verifier} \label{verifier} 
The Verifier module supervises the sequence of skills $\mu^{1:N}_t$ proposed by the Explorer, assessing whether the imagined transition is plausible, physically feasible, and stable.
Unlike the Explorer---which operates on current context---the Verifier considers a broader temporal window by accessing recent transitions $\{\mathcal{G}_{t-h}, \mu_{t-h}^{1:N}, \cdots, \mathcal{G}_{t-1}, \mu_{t-1}^{1:N}\}$. 
This history provides insight into the agent's prior actions and their outcomes, allowing the Verifier to make informed judgments grounded in embodied experience.

Given the current observation and proposed plan, the Verifier predicts the likely outcome and compares it with the Explorer’s intended goal graph. Beyond semantic alignment, it also performs stability checks---such as detecting precarious object placements, occlusions, or workspace clutter---that may compromise successful execution. If the proposed plan is unsafe or unlikely to succeed, the Verifier recommends corrective interventions, such as reordering skills, repositioning obstructing objects, or decluttering the workspace.

The Verifier module returns a structured feedback signal $f_t$ composed of:
\begin{itemize}
    \item a binary decision: ``Yes'' if the plan is executable or ``No'' otherwise;
    \item and, for ``No'' cases, an explanation detailing rejection reasons (e.g., infeasibility, instability, or deviation from the goal).
\end{itemize}

\begin{examplebox}
\footnotesize
\textbf{Continuing Example.}
In our tabletop scenario, if the Explorer suggests stacking a cup on a tray that is already full, the Verifier may reject the plan due to instability: It may return:
\[
\begin{aligned}
    f_t = \text{No: Cannot place cup on tray --- unstable configuration. Suggest removing the Blue block } \\
    \text{from the Tray first, then placing the Red cup on the tray.}
\end{aligned}
\]
\end{examplebox}

\subsection{Memory Retrieval}\label{memory}
To support novelty-based exploration and informed decision-making, \algoname maintains a dynamic memory module $\mathcal{M}$ that stores previously encountered scene graphs $\mathcal{G}$ derived from past observations. These structured graphs serve as compact, symbolic summaries of prior interactions, enabling the system to reason over what has already been seen and done.

Each $\mathcal{G}$ is instantiated using the NetworkX library, allowing efficient graph storage, manipulation, and querying.
At each timestep $t$, the Explorer queries $\mathcal{M}$ to retrieve a set of past scene graphs that are structurally similar to the current scene graph $\mathcal{G}_t$.
Specifically, the retrieved set is defined as 
\begin{equation}
    \{ \mathcal{G}_j \in \mathcal{M} \mid \text{dist}(\mathcal{G}_t, \mathcal{G}_j) < \tau \}, 
\end{equation}
where $\text{dist}(\mathcal{G}_t,\cdot)$ denotes an edit-based graph distance from the current scene $\mathcal{G}_t$, and $\tau$ is a predefined similarity threshold. This retrieval process surfaces relevant prior experiences that guide the Explorer in imagining transitions that are both novel (i.e., not previously observed) and physically plausible given historical outcomes.

In parallel, the Verifier leverages the same memory to assess the feasibility of proposed skill sequences, using previously executed transitions as empirical priors for prediction. Thus, memory plays a \textbf{dual role}: enabling semantic novelty in the imagination process and providing a grounded context for physical verification.

\begin{examplebox}
\footnotesize
\textbf{Continuing Example.}
Suppose the current scene $\mathcal{G}_t$ encodes a cup next to a tray, with a block positioned above.
The memory might retrieve a prior scene where the block was stacked on the cup and the tray was empty, helping the Explorer avoid redundant imagination and providing the Verifier with context on whether that configuration was previously successful or unstable.

\end{examplebox}

\begin{wrapfigure}{r}{0.4\linewidth}
    \centering
    \vspace{-0.2in}
    \includegraphics[width=\linewidth]{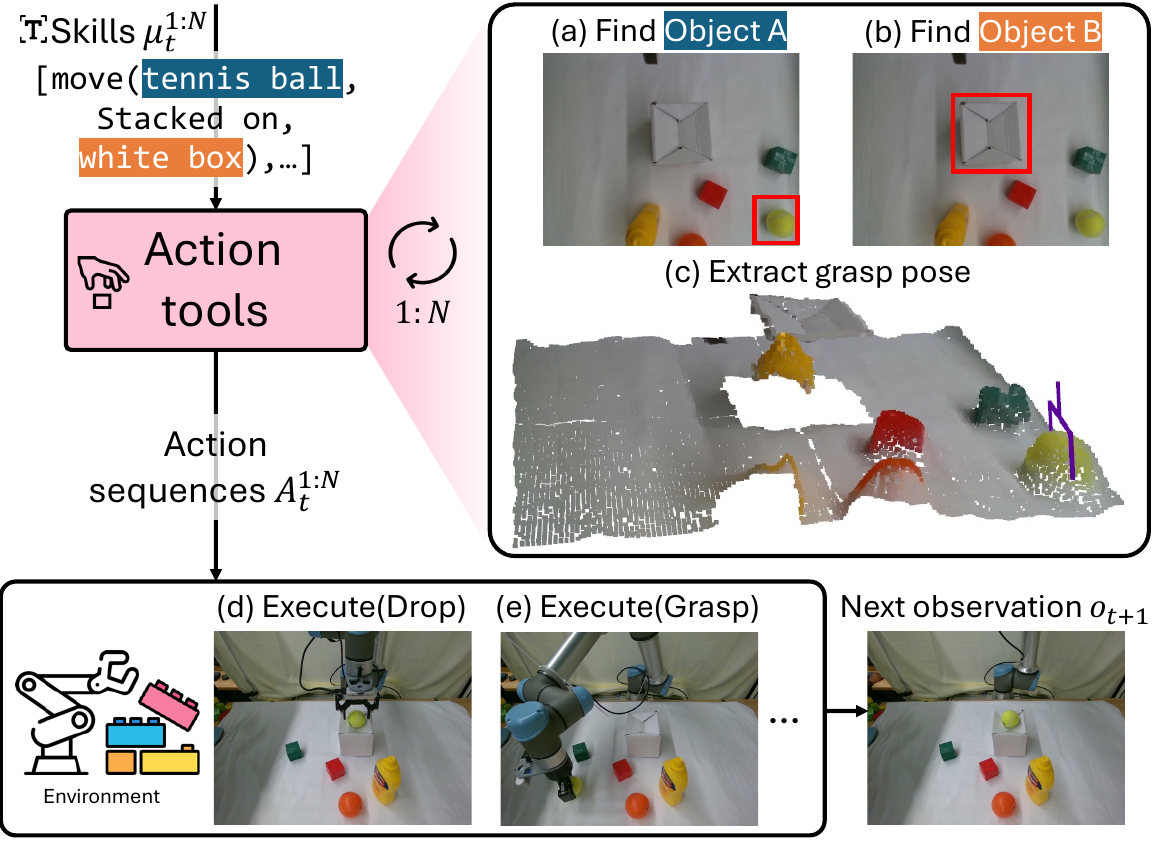}
    \caption{Example of transforming skill to action in a real-world environment. \label{fig:actiontool_eg}}
    \vspace{-0.2in}
\end{wrapfigure}

\newpage

\subsection{Action Tools}\label{action_tool}

To bridge the high-level skill sequence $\mu_t^{1:N}$ generated by the Explorer with real-world execution, we employ an \textit{Action Tools} module that translates each skill into a corresponding low-level action sequence $A_t^{1:N}$ using RGB-D observations of the current scene.
We first instantiate each skill $\mu_t^i$ with a predefined set of task-specific primitives that are aligned with the robot’s embodiment and operational constraints.
After executing a skill, \algoname records the resulting state, constructs the updated scene graph $\mathcal{G}_{t+1}$, and stores the transition tuple $(\mathcal{G}_t, \mu_t^{1:N}, \mathcal{G}_{t+1})$ in memory, completing a full imagination-verification-execution cycle. Details of the categories and modularities of the action tools are provided in the appendix \ref{action_tool_details}.

\section{Experiments}

Our experiments address three main questions. \textbf{Exploration quality}: How does \algoname compare to conventional exploration strategies in producing diverse and meaningful interactions (Figure~\ref{exploration_capability})? \textbf{Component effectiveness}: How does each module contribute to the overall performance of the system (Figure~\ref{ablation_study})? \textbf{Downstream utility}: How useful is the collected data for downstream tasks (Tables~\ref{tab:bc_success_rates},~\ref{tab:wm_success_rates})?

\subsection{Experimental Setup} 

\topic{Simulation Environment}
We use VimaBench \citep{jiang2023vima}, a simulated tabletop environment containing multiple rigid objects with varied shapes and colors. 
The robot is equipped with a suction end-effector and can execute pick-and-place in continuous action space.
To efficiently explore this environment, reasoning over spatial configurations while maintaining physical interaction constraints is required.

\topic{Real-World Environment}  
We use a 6-DoF UR5e robot arm with a parallel gripper to interact with tabletop objects. 
A VLM, GPT-4o~\citep{openai2024gpt4o}, observes the current scene and proposes high-level actions, which are then parsed and executed through a low-level controller. 
We predict the Grasp poses using AnyGrasp~\citep{fang2023anygrasp}, with target objects segmented via LangSAM \citep{medeiros2023langsam}, closely following the approaches introduced in \cite{liu2024dynamem, liu2024ok}. 
We apply a heuristic preference for top-down grasps for improved reliability. 
We compute drop positions using depth data and segmentation masks. 
For the ``Stacked on'' relationship, we first calculate the midpoint of the target object using the object's segmentation mask, then we calculate the drop height using the depth data for the object region.
After each execution, we collect the robot pose, task success status, and RGB data and add them to the experience buffer for downstream tasks.

\subsection{\algoname Outperforms RL and Human Baselines in Exploration Diversity
}\label{exploration_capability_comparison}
We compare \algoname against intrinsic motivation RL methods and three human-controlled strategies. For RL baselines, we implement novelty-based intrinsic reward approaches such as \textit{RND}~\citep{burda2018exploration} and \textit{RE3}~\citep{pmlr-v139-seo21a}. Human baselines include: \textit{Expert}, given explicit objectives and operating the robot via action tools; \textit{Novice}, interacting freely via action tools, without instructions; and \textit{Moved by hand}, directly manipulates objects without involving the robot arm.

\begin{figure}[!ht]
    \centering
    \includegraphics[width=0.95\linewidth]{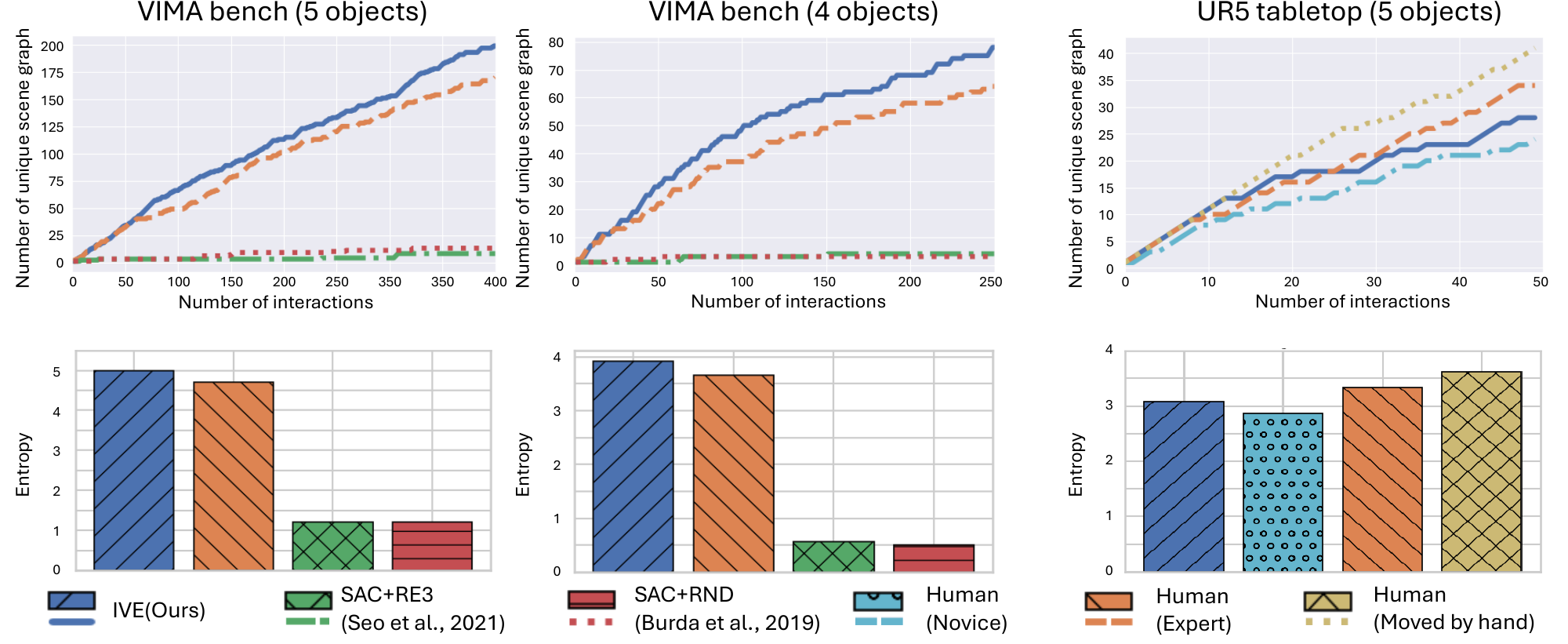}
    \caption{\textbf{Exploration capability evaluation across simulated and real-world environments.} Top: Growth curves of the number of unique scene graphs visited. Bottom: Diversity of visited states, measured by entropy (see Appendix~\ref{eexploration_metric} for the definition and computation details for entropy, and ~\ref{exploration_capability_comparison} for baseline details.}
    \label{exploration_capability}
\end{figure}

Figure \ref{exploration_capability} 
shows that 
\algoname achieves \textbf{4.1 to 7.8×} increase in state entropy over RL baselines and \textbf{82\% to 122\%} of the scene diversity exhibited by expert humans.
\footnote {Refer to the Appendix \ref{eexploration_metric} for details on how entropy and scene counts are computed.} 
We attribute this to humans gradually forgetting previously encountered scenes, whereas \algoname tracks all past configurations via retrieval. 
When considering only the first 50 transitions, human performance in simulation closely matches that of \algoname.
In the real world, \algoname slightly underperforms compared to a human expert. See Appendix \ref{other_vlms} for details on the comparative capabilities of different VLMs.

\topic{Why RL Falls Short} RL-based baselines perform worse than both \algoname and human participants. 
We believe this reflects a limitation of RL-driven exploration that prioritizes pixel-level novelty over semantically meaningful interactions, especially when they rely solely on intrinsic rewards, making structured and meaningful exploration challenging. In contrast, \algoname explores with semantic structure and memory-based recall, enabling higher-level diversity without external rewards.

\subsection{Ablations Reveal Memory, Explorer and Verifier Are Critical for Effective Exploration}

\begin{figure}[!ht]
    \centering
    \includegraphics[width=0.95\linewidth]{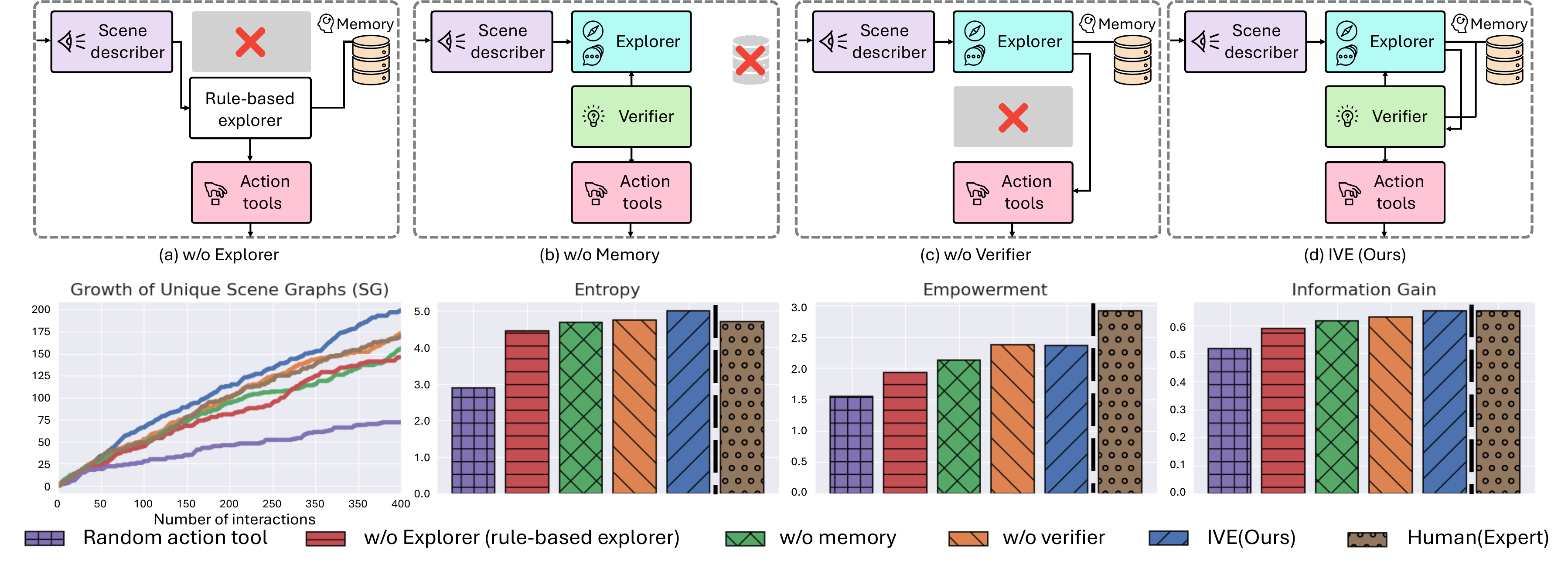}
    \caption{\textbf{Ablation study of \algoname.} 
    (Top) Illustration of each variant, highlighting removed modules in gray. 
    (Bottom) Exploration performance is measured by the number of unique scene graphs, entropy, empowerment, and information gain (see Appendix~\ref{eexploration_metric} for metric details). 
    }
    \label{ablation_study}
\end{figure}

We ablate components of \algoname to quantify their impact on exploration performance (Figure~\ref{ablation_study}). The following variants are tested:
\begin{itemize}[leftmargin=*]
    \item \textit{Random Tool Selector} uniformly samples action tools with no planning.
    \item \textit{w/o Explorer (Rule-Based Explorer)} uses simple rules to generate scene graphs; retains VLM for skill generation.
    \item \textit{w/o Memory} disables retrieval-based grounding.
    \item \textit{w/o Verifier} skips physical feasibility filtering.
\end{itemize}

We evaluate these variants based on the number of unique scene graphs visited, entropy, empowerment, and information gain. 
Figure \ref{ablation_study} shows that removing the memory module and the explorer results in a \textbf{22\% and 27\% drop in unique scenes discovered}, respectively, along with a reduction in entropy and information gain---highlighting its importance for novelty-aware planning. Verifier removal also degrades performance, although empowerment remains relatively stable---likely because the verifier proposes fewer decluttering actions, and such actions are inherently more stochastic and less goal-directed. 
The random tool selector baseline performs the worst, achieving less than half the number of unique graphs discovered by \algoname.

\begin{table}[htbp]
    \small
    \centering
    \caption{Performance of goal-conditional and non-conditional behavior cloning across tasks in simulation. Our method achieves human-level performance and significantly outperforms exploration RL baselines (RND~\citep{burda2018exploration} and RE3~\citep{pmlr-v139-seo21a}), demonstrating the effectiveness of our exploration strategy in generating diverse and semantically meaningful data. Example goal images used for the goal-conditioned tests are provided in Appendix~\ref{bc_evaluation}.}
    \label{tab:bc_success_rates}
    \resizebox{0.8\linewidth}{!}{
    \begin{tabular}{llccccc}
        \toprule
        & & \multicolumn{2}{c}{Non-conditional} & & & \multicolumn{1}{c}{Goal-conditional} \\
        \cmidrule(lr){3-7} %
        & Exploration Method & \# of achieved tasks & Entropy &  & & Success rate\\
        \midrule
        \multirow{4}{*}{VIMA Bench (5 objects)} 
        & SAC~\citep{haarnoja2018soft} + RND~\citep{burda2018exploration}   & 2.0 & 1.907 & & & 8.33\% \\
        & SAC~\citep{haarnoja2018soft}+ RE3~\citep{pmlr-v139-seo21a}   & 2.1 & 1.754 &  & &0.00\% \\
        & \algoname (Ours)  & \textbf{4.1} & \textbf{2.283} & &  &\textbf{58.33\%} \\
        & Human       & \underline{3.6} & \underline{2.021} & & & \underline{50.00\%} \\
        \midrule
        \multirow{4}{*}{VIMA Bench (4 objects)} 
        & SAC~\citep{haarnoja2018soft} + RND~\citep{burda2018exploration}   & 2.0 & 1.907 & & & 0.00\%  \\
        & SAC~\citep{haarnoja2018soft} + RE3~\citep{pmlr-v139-seo21a}   & 1.2 & \underline{1.959} & & & 0.00\%  \\
        & \algoname (Ours)  & \underline{3.1} & 1.528  & & & \textbf{41.67\%} \\
        & Human       & \textbf{4.2} & \textbf{1.897} & & & \underline{33.33\%}  \\
        \bottomrule
    \end{tabular}
    }
\end{table}

\begin{table}[htbp]
    \small
    \centering
    \caption{Quantitative evaluation of World Model (WM) predictions using datasets collected by different exploration methods, trained with DINO-WM \citep{zhou2024dino}.}
    \label{tab:wm_success_rates}
    \resizebox{0.9\linewidth}{!}{
    \begin{tabular}{lcccccc}
        \toprule
        \multicolumn{1}{c}{}  & \multicolumn{2}{c}{Sim Env 1} & \multicolumn{2}{c}{Sim Env 2} & \multicolumn{2}{c}{Real World}\\
        \cmidrule(lr){2-7}
        Exploration Method & SSIM (↑) & LPIPS (↓) & SSIM (↑) & LPIPS (↓) & SSIM (↑) & LPIPS (↓) \\
        \midrule
        SAC~\citep{haarnoja2018soft} + RND~\citep{burda2018exploration} & 0.812 ± 0.039 & 0.198 ± 0.060 & \underline{0.855 ± 0.036} & 0.168 ± 0.061 & - & - \\
        SAC~\citep{haarnoja2018soft} + RE3~\citep{pmlr-v139-seo21a} & 0.814 ± 0.040 & 0.199 ± 0.057 & 0.850 ± 0.034 & 0.168 ± 0.059 & - & - \\
        \algoname (Ours) & \textbf{0.837 ± 0.032} & \underline{0.129 ± 0.044} & 0.853 ± 0.032 & \underline{0.160 ± 0.058} & 0.634 ± 0.075 & \textbf{0.181 ± 0.056} \\
        Human & \underline{0.833 ± 0.032} & \textbf{0.126 ± 0.042} & \textbf{0.862 ± 0.027} & \textbf{0.139 ± 0.047} & \textbf{0.653 ± 0.072} & 0.194 ± 0.056 \\
        \bottomrule
    \end{tabular}
    }
\end{table}

\subsection{\algoname Enables Stronger Policy Learning and World Modeling}

We evaluate two downstream tasks to validate the usefulness of the data from the exploration.
Performance is compared across datasets collected by three strategies: our method, RL-based exploration (SAC~\citep{haarnoja2018soft} with RND~\citep{burda2018exploration} and RE3~\citep{pmlr-v139-seo21a}), and human demonstrations (WM only).

\begin{itemize}
\item \textbf{World Model Accuracy (WM)}: Measures the prediction accuracy of a learned dynamics model trained on the exploration data using DINO-WM \citep{zhou2024dino}. 
\item \textbf{Behavior Cloning (BC)}: Evaluates how well a Diffusion Policy \citep{chi2023diffusion}, trained on the exploration dataset, can achieve novel goals presented as images. 
\end{itemize}

As shown in Table~\ref{tab:bc_success_rates}, policies trained on \algoname data outperform those trained on RL exploration data by up to \textbf{+58\% in task success}, and achieve performance \textbf{on par with human demonstrations}. Similarly, in world model (WM) prediction tasks,~\Tref{tab:wm_success_rates} shows that \algoname achieves performance closest to that of human-collected data. 
The results on both policy learning and world model prediction highlight the benefits of its structured and diverse exploration strategy for downstream tasks.

\section{Conclusion}

We introduced \algoname, an agentic exploration framework that integrates imagination, verification, and execution to enable efficient exploration in robotic systems.
By leveraging the broad knowledge of VLMs, our method enables robots to explore and interact with their environment autonomously.
The imagine-verify-execute cycle in \algoname promotes high-level semantic diversity during exploration, resulting in rich datasets for learning downstream tasks.
In future work, we plan to expand the set of action tools available to the system, enabling more complex interactions and improving the generality of agentic exploration.


\clearpage

\topic{Limitations}
While our method performs well, it shares common limitations of real-world robotic systems. VLM-based reasoning introduces some latency, which could be reduced with lighter or distilled models. Our action tools are manually defined, which limits scalability for complex tasks; integrating learned policies as tools could address this limitation. Finally, our reliance on open-vocabulary object detection can lead to failures with occluded or novel objects—future work could incorporate multi-view perception or interactive discovery to improve robustness.

\acknowledgments{
This work was partially supported by (a) NSF CAREER Award (\#2238769) to AS, and (b) DARPA TIAMAT (\#80321), NSF Award (\#2147276 FAI), DOD AFOSR Award (\#FA9550-23-1-0048), and Adobe, Capital One and JP Morgan faculty fellowships to FH. The authors acknowledge UMD's supercomputing resources made available for conducting this research. The U.S. Government is authorized to reproduce and distribute reprints for Governmental purposes notwithstanding any copyright annotation thereon. The views and conclusions contained herein are those of the authors and should not be interpreted as necessarily representing the official policies or endorsements, either expressed or implied, of NSF, Adobe, Capital One, JP Morgan, or the U.S. Government. We thank Amir-Hossein Shahidzadeh, Eric Zhu, and Mara Levy for their help and advice.
}


\bibliography{example}  
\newpage
\appendix

\section{Evaluation Metrics for Exploration Capability}\label{eexploration_metric}

To quantitatively evaluate the agent's ability to explore its environment, we use the following metrics that capture diversity, informativeness, and control capacity over future states, as described in ~\cite{lidayan2025intrinsically}. We treat the state space $\mathcal{S}$ ($s \in \mathcal{S}$) as a discrete set of scene graphs to enable interpretable analysis.

\begin{itemize}
    \item \textbf{Unique Scene Graphs}: The number of distinct scene graphs encountered during exploration. A higher count reflects greater semantic diversity.
    
    \item \textbf{State Entropy}: Measures the entropy of the agent's state visitation distribution. Let $N_s$ denote the number of visits to state $s$. Then $p(s) = N_s / \sum_{s'} N_{s'}$, and entropy is given by:
    \[
    H(S) := -\sum_s p(s) \log p(s)
    \]

    \item \textbf{Information Gain (IG)}: Quantifies how much new information is acquired in each episode. Define $N_{s,a}^e$ as the count of action $a$ taken in state $s$ up to episode $e$. The information gain for episode $e$ is:
    \[
    IG^e := \frac{\sum_{(s,a)} IG_0^e(s,a)-IG_0^{e-1}(s,a)}{\sum_{(s,a)} N_{s,a}^e - N_{s,a}^{e-1}},
    \]where $IG_0^e(s,a):=\log(1+N_{s,a}^e)$.

    \item \textbf{Empowerment}: Captures the agent's control over future outcomes. Defined as the mutual information between actions and resulting states:
    \[
    E := \max_{p(a)} I(s'; a \mid s) = \max_{p(a)} \sum_{s',a} p(s'|a)p(a) \log \frac{p(s'|a)}{\sum_{a'} p(s'|a')p(a')}
    \]
    Since exact computation is intractable, we approximate $p(s'|a)$ via sampling and scene graph transition statistics.
\end{itemize}

For fair comparison, we quantize observations into scene graphs using methods that differ from those employed in \algoname.
In simulation, we construct scene graphs using a heuristic based on ground-truth object positions, encoding relative distances and spatial relationships.
In the real world, where ground-truth positions are unavailable, we generate scene graphs using a separate perception pipeline.
Importantly, both the prompt design and scene graph structure used for evaluation are distinct from those used in \algoname.
Please note that in both settings, none of the agents---including \algoname and all baselines---have access to the ground-truth object positions or the internal graphs used for evaluation.

\section{Real world robot setup}
The system is implemented on a Universal Robot UR5e robot arm with a Robotiq 2F-85 gripper. An Intel Realsense D435i depth camera is mounted on the robot end-effector. The workspace is a tabletop workspace with predefined boundaries. All robot poses are clipped to the workspace boundaries for safety.

\subsection{Grasp Planning and Execution}
The system implements two grasp strategies. The primary strategy uses Anygrasp~\citep{fang2023anygrasp} for grasp pose detection, which provides the grasp pose in the camera frame. We transform this to the robot base frame using the camera calibration and convert the \(4\times 4\) matrix to an axis-angle rotation vector using the Rodrigues method. The execution sequence moves to a clearance height \(0.1\text{m}\) above the grasp pose, aligns rotation, descends to the grasp pose with a z-offset of \(-0.048\text{m}\). The object is lifted to a clearance height to avoid collision and then moved to the destination pose.

The tangram grasp strategy uses a centroid-based approach. We compute the object centroid using image moments, project it to 3D using the pinhole camera model, and transform to the robot base frame. The grasp pose is set to the centroid position with a z-offset of \(-0.01\text{m}\) and a fixed rotation of \([0, -\pi, 0]\).

\subsection{Relation-based Placement and Tangram Manipulation}

The system implements six spatial relationships: STACKED ON, IN\_FRONT\_OF, BEHIND, TO\_LEFT\_OF, TO\_RIGHT\_OF, and ARRANGE. For each relationship, we compute the drop point using a relationship-specific algorithm.

\begin{itemize}
    \item STACKED ON relationship calculates the drop height by finding the maximum z-coordinate of the target object in the depth image and adding a drop height offset of \(0.01\text{m}\).
    \item IN\_FRONT\_OF and BEHIND relationships compute a y-offset of \(\pm0.08\text{m}\) from the target object's bounding box.
    \item TO\_LEFT\_OF and TO\_RIGHT\_OF relationships use an x-offset of \(\pm0.08\text{m}\).
    \item ARRANGE relationship employs a depth-based ground plane detection algorithm. We create a ground mask by thresholding the depth image at the table height, erode it using a kernel size proportional to the manipulated object's dimensions, and apply boundary constraints excluding regions very close to the workspace edges. From the valid placement regions, we randomly sample a placement point to introduce diversity while maintaining safety constraints.
\end{itemize}

For tangram manipulation, we implement a specialized edge alignment system that first detects polygon edges using contour detection with an epsilon ratio of 0.02. The system then compares the source and destination masks to find the optimal edge alignment. For each pair of edges, we compute the alignment angle by finding the angle between the edge vectors, considering both parallel and anti-parallel alignments. Since tangram pieces are constrained to rotate only in the z-axis, we compute the rotation matrix around the z-axis using the alignment angle. The translation is determined by computing the vector between the midpoints of the aligned edges, with additional jitter sampling to account for small variations in placement. We evaluate each potential alignment by computing the contact length between the edges and applying an occlusion penalty based on the overlap between the source and destination masks. The system selects the alignment with the maximum contact length while minimizing occlusion. All coordinate transformations between camera and robot frames are handled using standard eye-in-hand calibration and the pinhole camera model.

\subsection{Region-based Placement}

\begin{figure}[ht]
    \centering
    \includegraphics[width=0.4\linewidth]{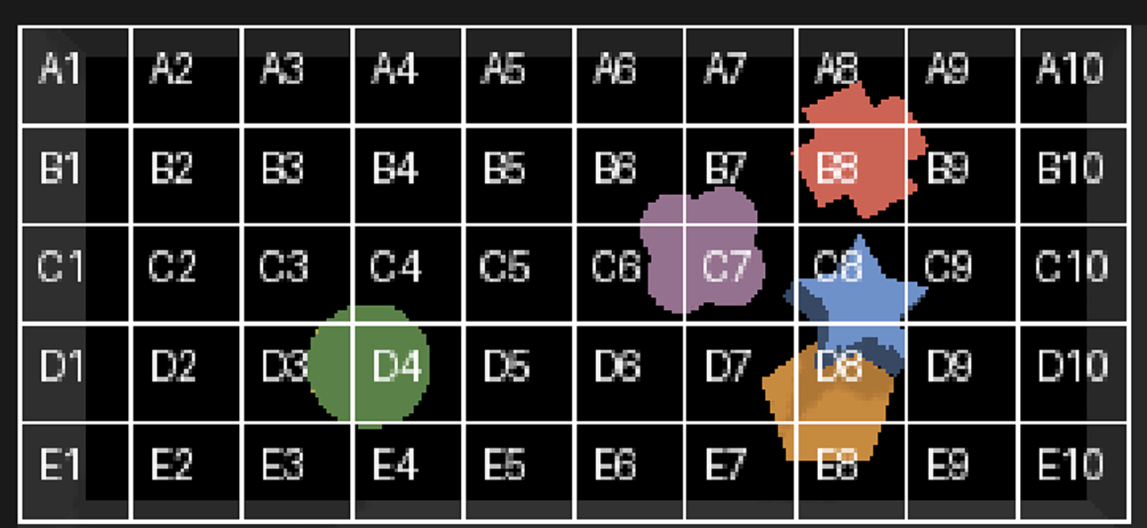}
    \caption{The Region-Based Placement Tool overlays the workspace with a labeled grid, allowing the VLM to reference specific spatial locations when issuing placement commands.}
    \label{fig:region_based}
\end{figure}

To enable the Vision-Language Model (VLM) to refer to specific spatial locations in the workspace, we introduce a Region-Based Placement Tool that discretizes the environment into a labeled grid map (Figure \ref{fig:region_based}). The workspace is overlaid with a checkerboard-style grid, where each cell is uniquely indexed (e.g., A1 to E10). This grid is rendered as an image and passed to the VLM.

Given this structured input, the VLM can issue explicit placement instructions using symbolic coordinates:
\texttt{move(object\_name, target\_grid)}. For instance, the command \texttt{move(red\_cross, B8)} indicates that the object referred to as \texttt{red\_cross} should be placed in cell \texttt{B8}. This discrete representation allows the model to generate unambiguous spatial commands and simplifies the mapping from language to robot actions.

\newpage
\section{Comparative Exploration Capabilities Across VLMs}\label{other_vlms}
In this section, we present the exploration performance of \algoname\ when paired with different types of Vision-Language Models (VLMs).
\begin{figure}[!ht]
    \centering
    \includegraphics[width=1\linewidth]{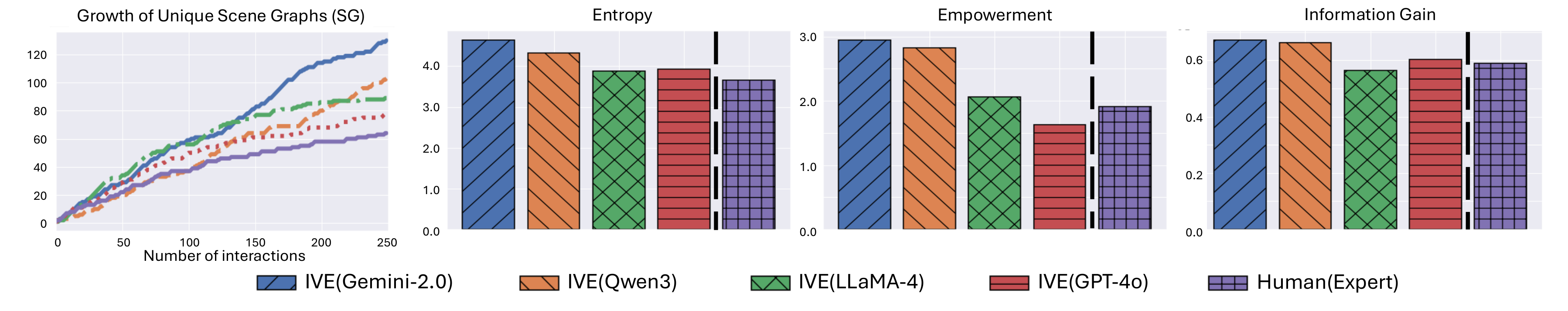}
    \caption{\textbf{Exploring with Embodied Agents:} This figure compares the exploration capabilities of our method, \algoname, powered by different Vision-Language Models (VLMs) rather than GPT-4o. Notably, \algoname, regardless of the VLM used, matches or surpasses the human expert in generating unique scene graphs, achieving higher state diversity, and gaining more information.}
    \label{fig:vlm_comparison}
\end{figure}

\section{Action Tool Details}\label{action_tool_details}

\paragraph{Tool Categories.}
Our action toolset includes three discrete types of manipulators:
\begin{enumerate}
\item \textbf{Relation-Based Placement Tools:} Execute relational actions that position objects with respect to others (e.g., ``To the left of,’’ ``Stacked on’’) as shown in Figure~\ref{fig:actiontool_eg}.
\item \textbf{Region-Based Placement Tools:} Place objects at specific locations on a predefined 2D layout (e.g., grid cells).
\item \textbf{Arranger Tool:} Manages workspace cleanliness by moving unreferenced or obstructive objects to free, uncovered regions, enabling subsequent actions.
\end{enumerate}

\paragraph{Modularity.}
The design of the Action Tools module is deliberately modular and extensible—new primitives or skills can be incorporated seamlessly without requiring architectural changes to other components. This modularity ensures that \algoname can adapt to diverse task domains and hardware platforms by \textit{swapping} or \textit{extending} action capabilities.


\section{Evaluations on Downstream Tasks}\label{bc_evaluation}

\begin{figure}[ht]
    \centering
    \includegraphics[width=1\linewidth]{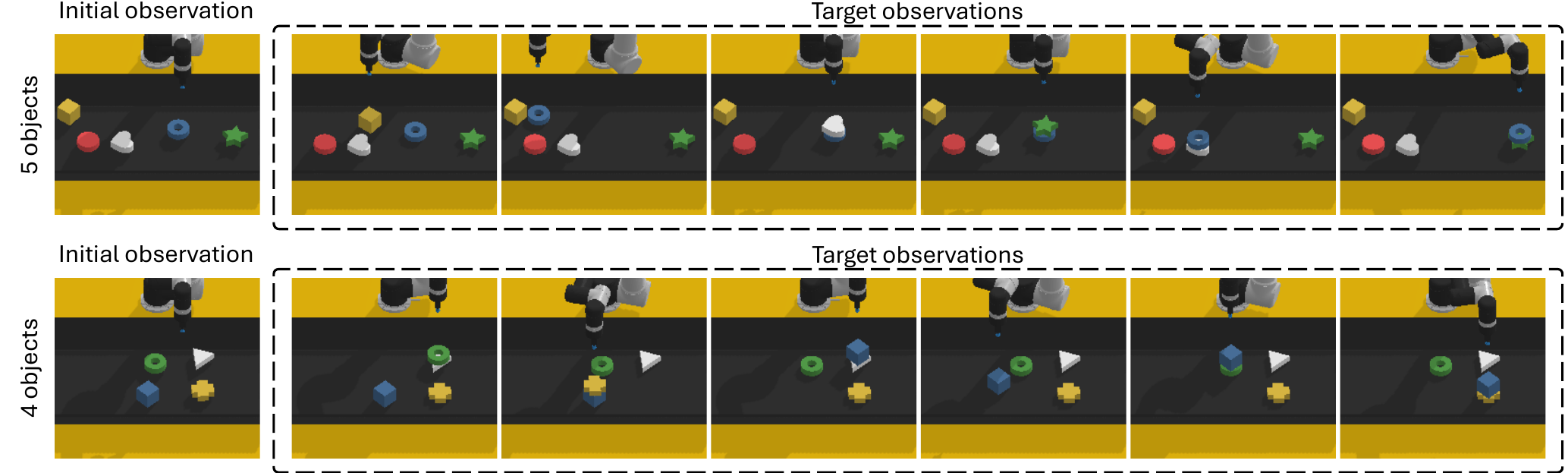}
    \caption{To evaluate the performance of the behavior cloning policy, we train Diffusion Policy \citep{chi2023diffusion} on each dataset and evaluate it on goal-conditioned tasks, where the initial observation is fixed and the agent is tested with six different goals.}
    \label{bc_goalcond_examples}
\end{figure}

\newpage

\begin{figure}[h!]
    \centering
    \includegraphics[width=0.93\linewidth]{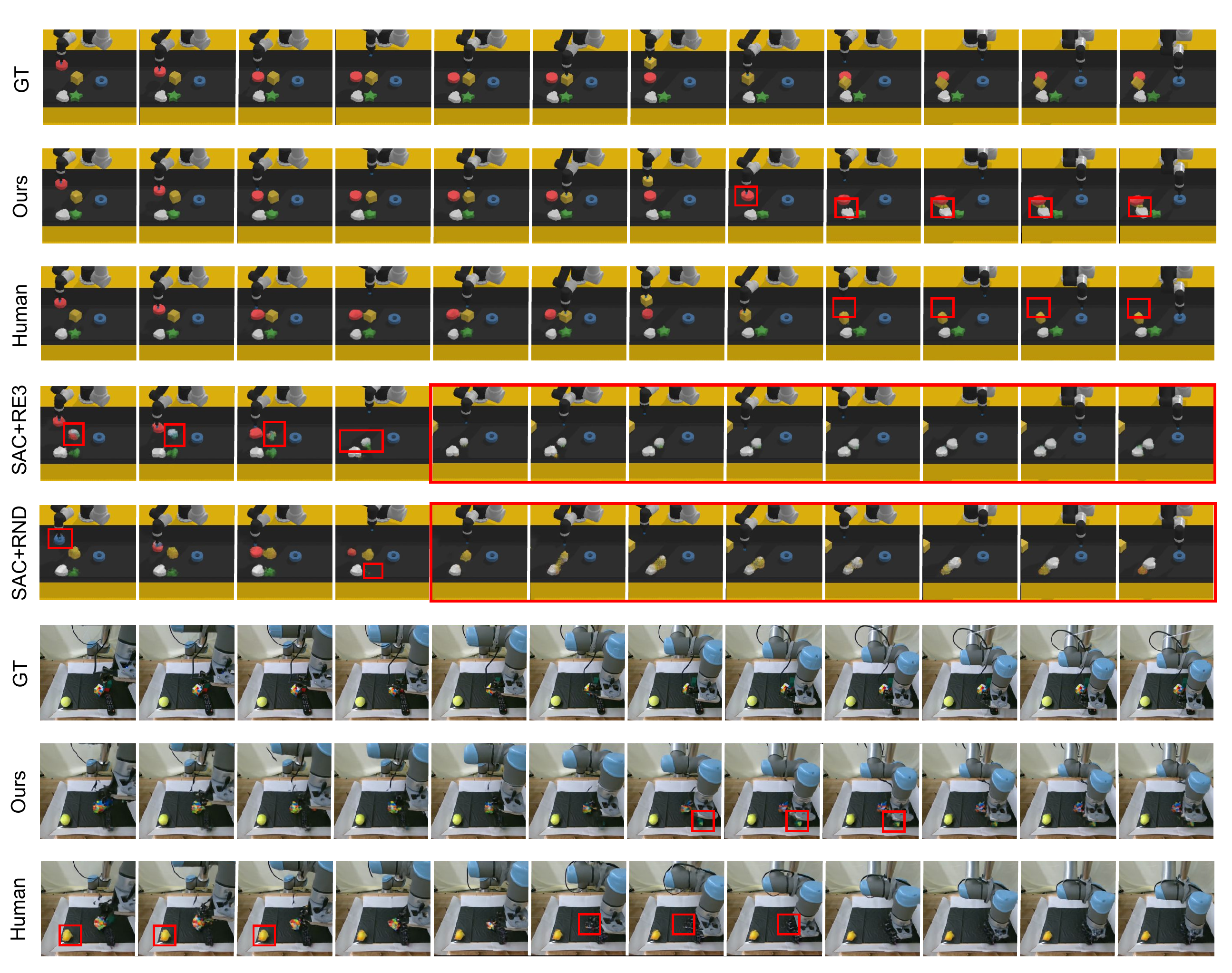}
    \caption{Qualitative examples of World Model (WM) predictions using datasets collected by different exploration methods. Red rectangles highlight regions with notable prediction errors.}
    \label{fig:enter-label}
\end{figure}

\section{Prompts for \algoname}
\subsection{Scene Describer}

The Scene Describer takes an RGB image and produces a structured scene graph that captures object identities and spatial relations. This module enables symbolic reasoning by abstracting the raw observation into a graph-structured representation. The prompt below guides a vision-language model to construct this graph iteratively.

The process consists of three main stages:

\begin{itemize}
  \item Step 1 - QnA Section: For each object, the model predicts its closest objects and describes their spatial relationship from that object's perspective.
  \item Step 2 - Iterative Scene Graph Construction: The scene graph is built incrementally by adding one object at a time and determining its relation to previously introduced objects.
  \item Step 3 - Final Scene Graph Output: The final graph is compiled from all previously gathered information, listing nodes (objects) and edges (relations) using only allowed relation types.
\end{itemize}

Below is the exact prompt used to guide the model:

\topic{Prompt}

\begin{lstlisting}
## Your Task
You are an expert image analyzer tasked with identifying the **exact** placement and spatial relationships of specific objects. Your job is to generate a scene graph describing these spatial relations **solely** based on the objects' visible positions in the image.

As an image analyzer, Follow Step 1~3 below.

---

## Step 1: Fill the Answer in QnA Section

---

## Step 2: Iterative Scene Graph Construction
1. Begin with one object.
2. Add one new object at a time, to your partial scene graph.
3. For each newly added object:
   - Determine its spatial relation(s) to the objects already in the scene graph.
   - **Use only** the Allowed Relations in the scene graph.
   - Do not assign more than one relation for the same object pair `(new_object, existing_object)' == `(existing_object, new_object)'
   - You may introduce multiple relations at once if the new object relates to multiple existing objects.

---

## Step 3: Final Scene Graph Output
1. **Once all objects** have been introduced and verified, compile a **complete scene graph**:
   - **List all nodes** (the objects in the final scene).
   - **List all verified relations** between pairs of objects, using the Allowed Relations in the scene graph.
2. **Use only** objects from the "Global Object Names."
3. Even if there's missing nodes or edges in a final scene graph (because at least one object is missing), you must still provide a complete **scratch pad** and **scene graph** with existing relations.

---

## Scene Graph Representation

- Nodes: Objects present in the scene.
- Relations: Spatial relationships between object pairs.
- Allowed Relations in the scene graph:
    - **Stacked On**: Object A is physically resting on Object B. This requires clear direct contact-Object A is visibly supported by Object B from below.
    - **Near**: Object A is positioned close to Object B without being stacked. Use this only when the objects are almost touching.

---
## Global Object Names
`<GLOBAL_OBJECTS_HERE>'
---
## Output Format
Please structure your final output exactly as shown below (without the lines). **Use the precise section titles**:
-------------
[Step 1: Fill the Answer in QnA Section]
<QNA_FOR_OBJECT_RELATION>
[Step 2: Iterative Scene Graph Construction]
Iteration 1:
- Added obj_a.
- Explanation of how you confirmed its presence in the image.
Iteration 2:
- Added obj_b.
- <obj_b, relation_type, obj_a> or <obj_a, relation_type, obj_b> (include any additional relations or notes)
- Explanation of how you verified this relation.
... (continue until all objects are added and checked)
[Step 3: Final Scene Graph Output]
<start_graph>
Nodes: obj_a, obj_b, ...
Relations: <obj_a, Near, obj_b>, <obj_b, Near obj_c>, <obj_d, Stacked On, obj_c>, ...
<end_graph>
-------------
\end{lstlisting}

\topic{QnA section}
Below is an example of a generated QnA section, filled out based on a set of sample object names:

\begin{lstlisting}
[Step 1: Fill the Answer in QnA Section]  
Object 1: red cube  
--------------------------------  
What are the closest 0~3 objects from red cube? What are their relations from red cube's perspective?  
Answer: The red cube is near the blue cylinder and stacked on the green base.  
--------------------------------  
Object 2: blue cylinder  
--------------------------------  
What are the closest 0~3 objects from blue cylinder? What are their relations from blue cylinder's perspective?  
Answer: The blue cylinder is near the red cube.  
--------------------------------  
Object 3: green base  
--------------------------------  
What are the closest 0~3 objects from green base? What are their relations from green base's perspective?  
Answer: The green base has the red cube stacked on top.  
--------------------------------  
\end{lstlisting}

\subsection{Explorer}

The Scene Explorer module performs planning over an environment with objects. It receives a current scene graph and predict a valid action sequence that results in a novel configuration. This task challenges the model to reason about physics, constraints, and symbolic novelty.

The prompt includes:

\begin{itemize}
  \item Action history: Provides the model with previously executed action sequence.
  \item Scene graph history: Supplies the model with previously visited scene graphs (which is retrived from memory), encouraging novelty.
\end{itemize}

\begin{lstlisting}
## Your Task
You are an expert spatial planner. Given the Current Image, your job is to generate a sequence of actions that discover a new scene configuration-one that has not been seen before.
- In addition to the action sequence, you must provide the predicted future scene graph (desired scene graph) that results from these actions.
- You have two images taken from different camera viewpoints.
- You should provide at most `<NUM_STEPS_HERE>' actions.
---
## Scene Graph Representation
- Nodes: Objects present in the scene.  
- Relations: Spatial relationships between object pairs.  
- Allowed Relations in the scene graph:  
    - **Stacked On**: Object A is physically resting on Object B. This requires clear direct contact-Object A is visibly supported by Object B from below.
    - **Near**: Object A is positioned close to Object B without being stacked. Use this only when the objects are almost touching.
---
## Global Object Names
`<GLOBAL_OBJECTS_HERE>'
---
<ACTION_TYPES>
---
## Current Scene Graph
`<CURRENT_SCENE_GRAPH>'
---
## Scene Graph History
Shows previously visited scene graphs most similar to your current scene.
<SCENEGRAPH_HISTORY>
---
## Action History
`<ACTION_HISTORY>'
---
## Output Format
Your output format should look exactly like the content between the `-----'. **Do not** number the actions. It's important to wrap the action sequence between `<start_action_sequence>' and `<end_action_sequence>'. Also, write down the predicted future scene graph (desired scene graph - the final arrangement after all actions) between `<start_graph>' and `<end_graph>'.
-----
<start_scratch_pad>
Explain your reasoning:
- Why this is a novel scene
- Why the action sequence makes sense
- If there were oddities or contradictions in the histories, how did you account for possible collisions, suction errors, or clutter?
<end_scratch_pad>
Predict (Desired) Future Scene Graph:
<start_desired_scene_graph>
Nodes: obj_a, obj_b, ...
Relations: <obj_a, Near, obj_b>, <obj_b, Near obj_c>, <obj_d, Stacked On, obj_c>, ...
<end_desired_scene_graph>
Next Action Sequence:
<start_action_sequence>
<ACTION_SEQUENCE_EXAMPLE>
<end_action_sequence>
-----
### Important Considerations
1. Order Matters: Plan your actions so that preconditions are satisfied before you move an object.
2. Scene Boundaries: If an object is near the scene boundary, avoid pushing it further toward the edge or placing new objects in a risky position.
3. Manipulation (Suction) Constraints: 
   - The suction can only reliably pick the topmost exposed surface.
   - In cluttered areas, an attempt to move one object may cause unintended collisions or shifts in neighboring objects.
   - Stacking another object on top of an unstable object can lead to the object toppling over.
4. Note: The list of allowed relations in Action Types and the relations used in Scene Graph Representation ([Stacked On, Near]) may differ. Desired Scene Graph should use relations among <SCENEGRAPH_RELATIONS> only, same as other Scene Graphs. Please keep this in mind when planning your actions.
\end{lstlisting}

\topic{Action types}

Actions available to the scene explorer fall into the following categories. These are symbolic commands grounded in real-world physical execution, and the model may extend this vocabulary when necessary.

\begin{lstlisting}
### Action Types

Actions are formatted in two ways:

1. `move(obj_a, RELATION, obj_b)'
   e.g., `move(white cup, Stacked On, red plate)'
   - Moves one object to a position relative to another.
   - Allowed RELATION list: `[In Front Of, Behind, To The Left Of, To The Right Of, Stacked On]'


2. `move(obj_a, GRID_ID)'
   e.g., `move(blue ball, B3)'
   - Moves an object to a grid location on the image. (`["A1", "B3", ..., "E10"]')

3, `arrange(obj_a)'
   e.g., `arrange(red block)'
   - Pick up the objects and organize them in a clear area on the Workspace.
\end{lstlisting}

\subsection{Verifier}

The Scene Verifier is responsible for checking the validity and physical feasibility of a proposed action sequence in dynamic environments. It assesses whether the actions, when executed from the current scene, would produce the desired result without causing instability or unintended configurations.

One core component of this process is the transition history, a temporally ordered trace of the environment, alternating between scene graphs and actions:
\begin{quote}
Scene Graph$_{1}$ → Action$_1$ → Scene Graph$_{2}$ → Action$_2$ → ... → Scene Graph$_{n}$
\end{quote}
This history provides concrete grounding to reason about object configurations and action effects, enabling the verifier to anticipate unintended side-effects like toppling, occlusions, or manipulation errors.

The verifier simulates outcomes, checks for physical plausibility, and may provide targeted suggestions or recommend a decluttering strategy in edge cases.

\begin{lstlisting}
## Your Task
You are a spatial reasoning expert responsible for **verifying action plans** in physically dynamic environments.  
You ensure that a proposed sequence of actions logically leads from the current state to the desired scene graph, without triggering unintended outcomes.
You may also provide **targeted suggestions** or, in rare but necessary cases, recommend a **temporary shift to a decluttering strategy**.
---
## Goals
Given the current image (from two camera views), transition history, desired scene graph, and a proposed action sequence:
1. **Simulate** the effect of the action sequence from the current scene  
2. **Predict** the resulting scene graph  
3. **Compare** the predicted graph with the desired one  
4. **Evaluate physical feasibility and execution stability**  
5. **Provide a judgment**:
   - Valid and feasible
   - Invalid (with reason)
   - Valid but risky (suggest a targeted fix)
   - Too unstable to proceed (recommend declutter mode)
---
<ACTION_TYPES>
---
## Transition History
A sequence of alternating scene graphs and actions showing the environment's evolution.
`<TRANSITION_HISTORY>'
---
## Output Format
-----
<start_scratch_pad>
Step-by-step analysis:
- Simulate and predict the resulting scene graph.
Scene Stability Check:
- Are any objects in clearly unstable or unreachable positions?
- Do previous transitions indicate failures or ambiguous changes?
- Are cluttered zones, deep stacks, or occlusions affecting safety or reliability?
Decision:
- Is the action sequence logically valid and does it produce the desired scene graph?
  -> YES or NO
If NO:
- Explain which actions fail and why.
- Point out mismatches or invalid transitions.
If issues are detected:
- Identify objects or areas causing risk (e.g., unstable stacks, blocked objects).
- Suggest fine-grained intervention (e.g., "move obj_A before continuing").
If the environment is severely cluttered and unsafe:
- Recommend a temporary shift to a decluttering mode
<end_scratch_pad>
<start_decision>
YES or NO
<end_decision>
<start_reason>
[If NO: Brief but clear explanation of what failed or was mismatched]
- risky: Warning message with suggestion, e.g., "Unstable stack: move obj_b before continuing"
- Too unstable: "Scene too cluttered. Recommend temporary declutter mode."
[If YES and no issues: Leave this part empty]
<end_reason>
-----
---
## Scene Stability Considerations
Clutter or instability **does not always require full decluttering**. Consider recommending targeted fixes first.
#### Examples of Minor Intervention:
- `"obj_b is stacked on obj_a, which is already supporting obj_c. Recommend moving obj_b first to prevent instability."'
- `"obj_d is partially occluded and may be hard to suction. Recommend shifting nearby obj_e first."'
#### Examples of Decluttering (rare):
- `"Multiple overlapping clusters and deep stacks suggest high instability. Recommend decluttering of current layout before further scene exploration."'
\end{lstlisting}

\end{document}